\documentclass[]{template}

\usepackage[utf8]{inputenc}             
\usepackage[T1]{fontenc}                
\usepackage{url}                        
\usepackage{booktabs}                   
\usepackage{multirow}
\usepackage{colortbl}
\usepackage{multicol}
\usepackage{amsfonts}                   
\usepackage{nicefrac}                   
\usepackage{microtype}                  
\usepackage[dvipsnames]{xcolor}         

\usepackage{latexsym}

\usepackage{graphicx}
\usepackage{float}
\usepackage{subcaption}
\usepackage{wrapfig}
\usepackage{lipsum}
\usepackage{tikz}
\usetikzlibrary{arrows.meta,calc}

\usepackage[most]{tcolorbox}
\usepackage{xcolor}
\usepackage[section]{placeins}
\definecolor{takeawaybg}{RGB}{244,247,252} 

\newtcolorbox{takeawaysbox}{
  enhanced,
  breakable,
  colback=takeawaybg,
  colframe=black,
  boxrule=1.1pt,   
  arc=4pt,         
  left=10pt,
  right=10pt,
  top=7pt,
  bottom=7pt,
  boxsep=0pt,
  before skip=6pt,
  after skip=6pt
}

\usepackage{bm}

\usepackage{tabularx} 
\usepackage{ragged2e} 
\newcolumntype{L}{>{\RaggedRight\hangafter=1\hangindent=0em}X}

\usepackage{enumitem}

\usepackage{amsmath}
\usepackage{amssymb}
\usepackage{mathtools}
\usepackage{amsthm}

\setboolean{logo}{true}    

\usepackage[linesnumbered,ruled,vlined]{algorithm2e}
\usepackage{amsmath}
\usepackage{amssymb}
\usepackage{algorithm}
\usepackage{algpseudocode}

\hypersetup{
    colorlinks=true,
    linkcolor=red,
    citecolor=Cerulean,
    filecolor=magenta,      
    urlcolor=magenta,
}

\usepackage[capitalize,noabbrev]{cleveref}
\crefname{section}{§}{§§}
\Crefname{section}{§}{§§}

\usepackage{calligra}
\DeclareMathAlphabet{\mathcalligra}{T1}{calligra}{m}{n}

\usepackage{pifont}

\theoremstyle{plain}

\theoremstyle{definition}

\theoremstyle{remark}

\renewcommand{\paragraph}[1]{\vspace{1mm}\noindent\textbf{#1}}

\DeclareCaptionLabelFormat{cont}{#1~#2\alph{ContinuedFloat}}
\usepackage[most]{tcolorbox}
\tcbset{
  promptbox/.style={
    top=10pt,
    colback=lightgray!20,
    colframe=Black,
    colbacktitle=NavyBlue,
    enhanced,
    center,
    attach boxed title to top center={yshift=-0.1in,xshift=0.0in},
    boxed title style={boxrule=0pt,colframe=white,},
  }
}
\newtcolorbox{promptbox}[2][]{promptbox, title=#2,#1}
\tcbset{
  takeawaybox/.style={
    top=10pt,
    colback=lightgray!20,
    colframe=Black,
    colbacktitle=BurntOrange,
    enhanced,
    center,
    attach boxed title to top center={yshift=-0.1in,xshift=0.0in},
    boxed title style={boxrule=0pt,colframe=white,},
  }
}
\newtcolorbox{takeawaybox}[2][]{takeawaybox, title=#2,#1}
\tcbset{
  observationbox/.style={
    top=10pt,
    colback=lightgray!20,
    colframe=Black,
    colbacktitle=YellowGreen,
    enhanced,
    center,
    attach boxed title to top center={yshift=-0.1in,xshift=0.0in},
    boxed title style={boxrule=0pt,colframe=white,},
  }
}
\newtcolorbox{observationbox}[2][]{observationbox, title=#2,#1}

\usepackage{xspace}

\newcommand\blfootnote[1]{%
  \begingroup
  \renewcommand\thefootnote{}\footnote{#1}%
  \addtocounter{footnote}{-1}%
  \endgroup
}

\usepackage{CJK}

\title{Mousse: Rectifying the Geometry of Muon with Curvature-Aware Preconditioning}
\renewcommand{\today}{2026-03-11}

\author[1, 2]{Yechen Zhang}
\author[2, 3]{Shuhao Xing}
\author[2, 1]{Junhao Huang}
\author[3]{Kai Lv}
\author[2]{Yunhua Zhou}
\author[3]{Xipeng Qiu}
\correspondauthor[2]{Qipeng Guo}
\author[2]{Kai Chen}

\affil[1]{Shanghai Jiao Tong University}
\affil[2]{Shanghai AI Laboratory}
\affil[3]{Fudan University}

\begin{abstract}
Recent advances in spectral optimization, notably Muon, have demonstrated that constraining update steps to the Stiefel manifold can significantly accelerate training and improve generalization. However, Muon implicitly assumes an isotropic optimization landscape, enforcing a uniform spectral update norm across all eigen-directions. We argue that this "egalitarian" constraint is suboptimal for Deep Neural Networks, where the curvature spectrum is known to be highly heavy-tailed and ill-conditioned. In such landscapes, Muon risks amplifying instabilities in high-curvature directions while limiting necessary progress in flat directions. In this work, we propose \textbf{Mousse} (\textbf{M}uon \textbf{O}ptimization \textbf{U}tilizing \textbf{S}hampoo's \textbf{S}tructural \textbf{E}stimation), a novel optimizer that reconciles the structural stability of spectral methods with the geometric adaptivity of second-order preconditioning. Instead of applying Newton-Schulz orthogonalization directly to the momentum matrix, Mousse operates in a whitened coordinate system induced by Kronecker-factored statistics (derived from Shampoo). Mathematically, we formulate Mousse as the solution to a spectral steepest descent problem constrained by an anisotropic trust region, where the optimal update is derived via the polar decomposition of the whitened gradient. Empirical results across language models ranging from 160M to 800M parameters demonstrate that Mousse consistently outperforms Muon, achieving around $\sim$12\% reduction in training steps with negligible computational overhead.

\end{abstract}

\begin{document}

\maketitle

\section{Introduction}

\begin{figure}[!b]
  \centering
  \includegraphics[width=0.8\columnwidth]{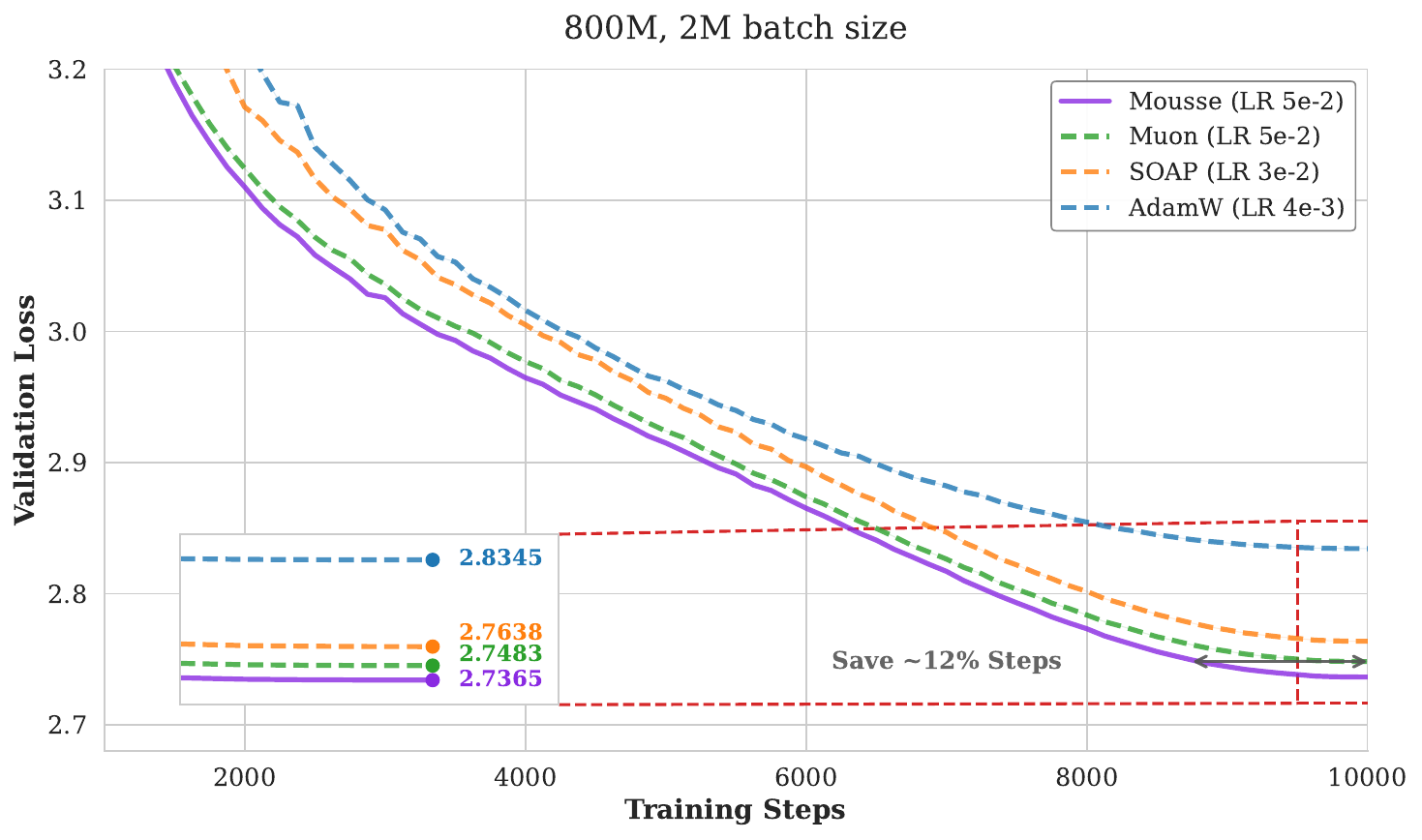}
  \caption{The optimal result of \textit{Muon} and \textit{Mousse} optimizers on 800M models. \textit{Mousse} achieves a $\sim$12\% reduction in training steps to reach comparable loss levels against \textit{Muon}.}
  \label{fig:experiments}
\end{figure}

Optimizing large language models requires navigating a high-dimensional and notoriously ill-conditioned loss landscape~\citep{sagun2016eigenvalues}. While adaptive optimization has firmly formed the backbone of deep learning --- spanning from the ubiquitous coordinate-wise AdamW~\citep{loshchilov2017decoupled} to advanced structural preconditioners like Shampoo~\citep{gupta2018shampoo} and SOAP~\citep{vyas2024soap} --- these methods primarily focus on refining the gradient direction within Euclidean geometry. Recently, a new paradigm of spectral optimization has gained significant traction in both academia and industry. Notably, the Muon optimizer~\citep{jordan2024muon} has emerged as a compelling alternative, constraining updates to the Stiefel manifold by functionally enforcing global spectral regularization via Newton-Schulz iterations. Leading AI laboratories, including Moonshot-AI~\citep{team2025kimi}, DeepSeek-AI~\citep{cheng2026conditional} and ZhipuAI~\citep{zeng2025glm}, have integrated Muon optimizer into their pre-training pipelines to capitalize on its superior scale-invariance and convergence efficiency at scale.

Despite its growing popularity, the efficacy of standard Muon is predicated on a stringent geometric assumption: it imposes an isotropic trust region, enforcing a uniform spectral norm across all eigen-directions. This "egalitarian" constraint treats all dimensions as geometrically equivalent, ignoring the vast disparities in curvature inherent to neural networks. Although recent community efforts have attempted to refine Muon in various ways~\citep{si2025adamuon}, the critical challenge of aligning Muon's isotropic constraint with the highly anisotropic curvature of neural network landscapes remains insufficiently explored.

To resolve this geometric discord, we introduce \textbf{Mousse} (\textbf{M}uon \textbf{O}ptimization \textbf{U}tilizing \textbf{S}hampoo's \textbf{S}tructural \textbf{E}stimation), a unified optimizer that synthesizes second-order preconditioning with spectral constraints. Our approach stems from a fundamental insight: the isotropic assumption of spectral updates is mathematically optimal only when applied within a spatially whitened geometry. Mousse achieves this by performing a change of basis: it first preconditions the gradient using Shampoo's Kronecker-factored curvature statistics \citep{gupta2018shampoo}, effectively "sphering" the local optimization landscape. Subsequently, the Newton-Schulz orthogonalization is applied in this transformed coordinate system. This procedure aligns the optimizer's rigid spectral constraint with the actual geometry of the loss surface, thereby retaining the stability of Stiefel manifold updates while accurately respecting the ill-conditioned curvature of deep neural networks. Our empirical results rigorously confirm that Mousse's performance gains are robust to scheduling choices and reflect a genuine geometric advantage. Specifically, Mousse reduces the training steps required to reach a target loss by approximately 12\%, while incurring only a 3\% wall-clock time overhead compared to standard Muon (see Figure~\ref{fig:experiments}).

Our contributions are summarized as follows:

\begin{itemize}
\item \textbf{A Unified Geometric Framework:} We theoretically ground Mousse as the optimal solution to the dual-norm maximization problem under anisotropic geometry. By reformulating spectral optimization within a whitened coordinate system derived from Kronecker-factored curvature statistics, Mousse rectifies the optimization landscape, bridging the gap between spectral methods and second-order preconditioners.

\item \textbf{Robust Engineering Insights:} We provide a comprehensive analysis of the stability challenges in combining spectral constraints with heavy-tailed curvature estimation. We introduce critical techniques such as \textbf{Trace Normalization} and \textbf{Spectral Tempering}, offering practical guidelines for stabilizing second-order spectral optimization in various training settings.

\item \textbf{Pareto-Optimal Efficiency:} Through extensive experiments on language models ranging from 160M to 800M parameters, we demonstrate that Mousse consistently outperforms Muon. By delivering substantial gains in sample efficiency that strictly outweigh the marginal computational cost, Mousse establishes a new state-of-the-art trade-off for large-scale pre-training.
\end{itemize}

\section{Related Works}
\blfootnote{Source code is available at \url{https://github.com/Anti-Entrophic/Mousse}.}
\paragraph{Adaptive and Second-Order Optimization.}
Standard adaptive methods like AdamW rely on element-wise heuristics, scaling updates based on diagonal Hessian approximations. To capture the rich parameter correlations ignored by these coordinate-wise methods, second-order optimizers such as K-FAC~\citep{martens2015optimizing} and Shampoo~\citep{gupta2018shampoo} utilize Kronecker-factored curvature statistics for preconditioning. SOAP further advances this direction by integrating Shampoo's structural preconditioning with Adam-style momentum and adaptive step sizes to enhance optimization performance, though it retains the computational burden of high-rank matrix operations. Muon~\citep{jordan2024muon} introduces strict spectral constraints to the parameter updates, revealing a highly promising new paradigm for efficient training. It is exactly along this promising trajectory that our proposed method, Mousse, explores the integration of advanced preconditioning with spectral efficiency.

\paragraph{Advances in Spectral Optimization.}
Following the rapid rise of spectral methods, a growing body of work has sought to analyze and refine the geometric constraints of optimization. Bernstein~\citep{bernstein2024old} notes that Shampoo mathematically reduces to Muon in the absence of memory, framing the latter as an instantaneous curvature correction. PolarGrad~\citep{lau2025polargrad} argues that standard spectral updates discard gradient magnitude information, proposing instead that updates should be scaled by the nuclear norm to retain intensity. Similarly addressing scale imbalances at the neuron level, NorMuon~\citep{li2025normuon} combines orthogonalization with row-wise adaptive learning rates to prevent specific neurons from dominating the update process. SUMO~\citep{refael2025sumo} restricts spectral decomposition to a dynamically adapted low-rank subspace, thereby aligning optimization steps with the principal components of the loss landscape. On the efficiency front, LiMuon~\citep{huang2025limuon} incorporates randomized SVD and variance reduction to improve memory usage and convergence guarantees. Mousse focuses fundamentally on the core spectral constraints of Muon. By explicitly introducing historical curvature into this spectral framework, Mousse is expected to achieve more precise and stable parameter updates.

\section{Methods}

In this section, we present the mathematical foundation of our method and introduce its concrete formulation by framing optimization as a geometric process connecting dual spaces. We denote the parameter space as $\mathcal{W}$ and the gradient space as $\mathcal{G} \cong \mathcal{T}_W^*\mathcal{W}$. The gradient $G = \nabla_W \mathcal{L}\in\mathcal{G}$ represents a linear functional on the tangent space $\mathcal{T}_W \mathcal{W}$,  and the role of an optimizer is to define a mapping $p: \mathcal{G} \rightarrow \mathcal{T}_W \mathcal{W}$ that translates this functional into a concrete update step $\Delta W$. We use $\text{vec}(\cdot)$ to denote the vectorization operator, which stacks the columns of a matrix into a single column vector.

\subsection{Muon as Spectral Steepest Descent}

Standard gradient descent algorithms can be unified under the framework of \textbf{steepest descent}, where the update step $\Delta W$ is chosen to minimize the local linear approximation of the loss $\mathcal{L}$ subject to a trust region constraint defined by a specific norm $\|\cdot\|$. Formally, the update is the solution to:
\begin{equation}
    \Delta W = \mathop{\mathrm{argmin}}_{U \in \mathcal{T}_W\mathcal{W}} \langle G, U \rangle \quad \text{s.t.} \quad \|U\| \leq 1
\end{equation}

Here, the choice of the norm $\|\cdot\|$ dictates the geometry of the optimization path. For SGD, the Frobenius norm ($\|\cdot\|_F$) implies an isotropic Euclidean geometry, leading to $\Delta W \propto -G$. For Adam-style optimizers, the norm approximates a coordinate-wise weighted $\ell_\infty$ metric.

Muon, however, adopts the spectral norm (denoted as $\| \cdot \|_{op}$), which measures the magnitude of a matrix by its largest singular value:
\begin{equation}
    \Delta W_{\text{Muon}} = \mathop{\mathrm{argmin}}_{U \in \mathcal{T}_W\mathcal{W}} \langle G, U\rangle, \quad \text{s.t.} \quad \|U\|_{op} \leq 1
\end{equation}

The exact solution to this problem is given by the polar decomposition of the negative gradient. In practice, Muon approximates this via the Newton-Schulz iteration, denoted as $\text{msign}(\cdot)$:
\begin{equation}
    \Delta W_{\text{Muon}} = -\text{msign}(G) = -U_{\text{svd}}V_{\text{svd}}^T
\end{equation}

\begin{figure}[t]
  \centering
  \includegraphics[width=0.8 \columnwidth]{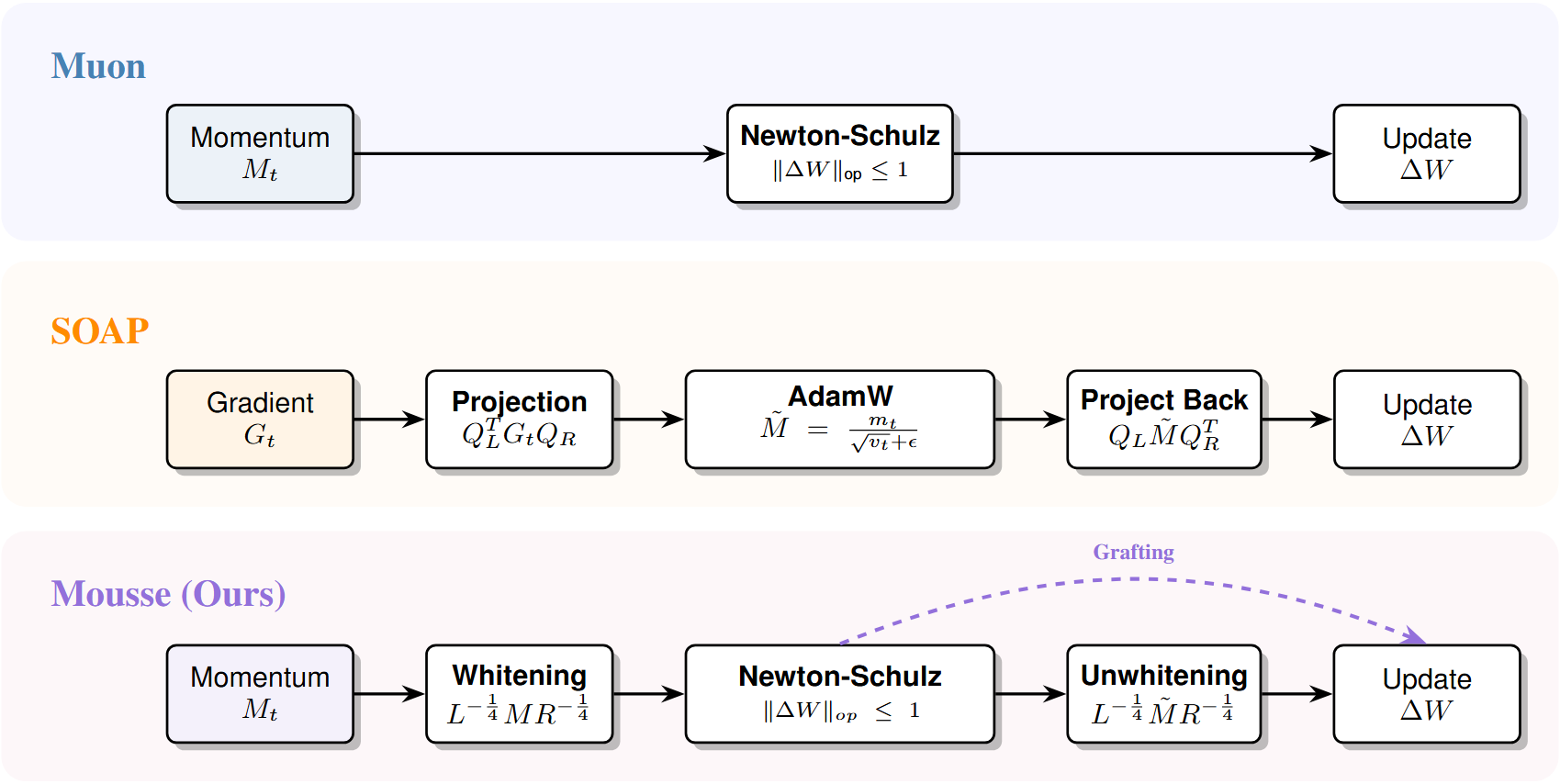}
  \caption{\centering Overview of the Mousse framework compared to baselines.}
  \label{fig:comparison}
\end{figure}

where $G = U_{\text{svd}}\Sigma V_{\text{svd}}^T$ is the singular value decomposition (SVD) of the gradient. This projection is equivalent to restricting the update onto the Stiefel manifold, ensuring $\Delta W^T \Delta W = I$. It is worth noting that the spectral norm provides a looser constraint than the Frobenius norm used in SGD, as $\|\cdot\|_{op} \leq \|\cdot\|_F$ holds true for any matrix. This implies that Muon permits updates with significantly larger total energy (under Frobenius norm) compared to SGD, while still strictly bounding the maximum step size along any single eigen-direction. This capacity for more aggressive updates without compromising directional stability is a key factor in its superior convergence.

\subsection{Mousse: The Geometric Framework}
\label{sec:mousse}

Despite these desirable properties of the spectral constraint, applying standard Muon directly to raw parameters is geometrically valid only if the local curvature is spherical. The steepest descent direction should therefore be defined with respect to this intrinsic geometry, rather than the extrinsic Euclidean coordinates.

To rectify this, Mousse reformulates the optimization problem by imposing the spectral constraint with regards to the anisotropic curvature. Let $H$ denote a positive semi-definite approximation to the local curvature. Then the size of an update should be measured not in the raw Euclidean coordinates, but under the quadratic form induced by $H$:
\begin{equation}
    \text{vec}(\Delta W)^T H \text{vec}(\Delta W) = C
\end{equation}

Equivalently, this geometry can be expressed in a whitened coordinate system. Defining
\begin{equation}
    \text{vec}(\tilde{U}) = H^{\frac{1}{2}} \text{vec}(U),
\end{equation}

the quadratic form becomes the standard Euclidean norm in the transformed space. Mousse then imposes Muon's spectral constraint in this whitened basis, yielding the following optimization problem:
\begin{equation}
    \Delta W = \mathop{\mathrm{argmin}}_{U \in \mathcal{T}_W\mathcal{W}} \langle G, U \rangle \quad  \text{s.t.}\,  \|\text{vec}^{-1}[H^{\frac{1}{2}}\text{vec}(U)]\|_{op} \leq 1
\end{equation}


Theoretically, different approximations of $H$ yield different algorithms. Mousse adopts Shampoo's \textbf{layerwise Kronecker-factored Hessian approximation} as the preconditioner, with $H \approx (R \otimes L)^{\frac{1}{2}}$, where $L$ and $R$ are estimated from the exponential moving averages of $GG^T$ and $G^T G$, respectively. This approximation has been shown to offer strong performance at comparable computational cost.
$$
\begin{aligned}
     L_t &= \frac{\beta_{pc}}{1 - \beta_{pc}^t} L_{t-1} + \frac{1-\beta_{pc}}{1 - \beta_{pc}^t} G_t G_t^T \\
     R_t &= \frac{\beta_{pc}}{1 - \beta_{pc}^t} R_{t-1} + \frac{1-\beta_{pc}}{1 - \beta_{pc}^t} G_t^T G_t 
\end{aligned}
$$

Under this structure, the whitening operator becomes a sandwich product
\begin{equation}
\tau_H(U) = (R\otimes L)^{\frac{1}{4}} \text{vec}(U) = L^{\frac{1}{4}}UR^{\frac{1}{4}}
\end{equation}

Let $P = L^{\frac{1}{4}}$ and $Q = R^{\frac{1}{4}}$ be the whitening factors for the rows and columns, respectively. The optimization objective of Mousse is formally:
\begin{equation}
    \min_{\Delta W} \text{Tr}(G^T \Delta W) \quad \text{s.t.} \quad \|P \Delta W Q\|_{\text{op}} \leq 1
\end{equation}

Here, the natural pairing between the gradient $G$ and the update $\Delta W$ is defined via the trace operator (See Appendix~\ref{appendix:natural_pairing}). Let $Y = P\Delta W Q$ represent the update step in the whitened coordinate system. Substituting $\Delta W = P^{-1} YQ^{-1}$ into the linear objective yields:
\begin{equation}
    \min \text{Tr}(G^TP^{-1}YQ^{-1}), \quad \text{s.t.} \|P\Delta W Q\|_{\text{op}} \leq 1
\end{equation}

Exploiting the cyclic property of the trace operator and the symmetry of the PSD matrices $P$ and $Q$, we can further rewrite the objective as:
$$
\begin{aligned}
    \text{Tr}(G^T P^{-1} Y Q^{-1}) &= \text{Tr}(Q^{-1}G^TP^{-1}Y) \\
    &= \text{Tr}([P^{-T}GQ^{-T}]^T Y) \\
    &= \text{Tr}([P^{-1}GQ^{-1}]^T Y) \\
\end{aligned}
$$

If we define $\tilde{G} = P^{-1}GQ^{-1}$ as the preconditioned gradient, then the problem now simplifies to the standard Muon form in terms of $Y$:
\begin{equation}
\min_{Y} \text{Tr}(\tilde{G}^TY), \quad \text{s.t.} \|Y\|_{\text{op}} \leq 1
\end{equation}

The corresponding solution is analogous to that of Muon:
$$
\begin{aligned}
Y &= -\text{msign}(\tilde{G}) \\
\Rightarrow \Delta W &= -L^{-\frac{1}{4}} \text{msign}(L^{-\frac{1}{4}}GR^{-\frac{1}{4}})R^{-\frac{1}{4}} \\
\end{aligned}
$$

\begin{algorithm}[!t]
\caption{Mousse Optimizer}
\label{alg:mousse}
\begin{algorithmic}[1]
\Require \textbf{Hyperparameters:} Learning rate $\eta$, Momentum coefficient $\beta$, Preconditioner moving average $\beta_{pc}$, Damping factor $\epsilon$, Curvature exponent $\alpha$, Preconditioner update interval $T$.
\Require \textbf{Initial States:} Momentum $M \leftarrow 0$, Parameters $\theta$.

\vspace{0.4em}
\State $M \leftarrow \beta M + G$

\vspace{0.4em}
\Statex \textbf{\textit{Update Curvature Statistics}}
\State $L \leftarrow \beta_{pc} L + (1-\beta_{pc}) G G^T$, $R \leftarrow \beta_{pc} R + (1-\beta_{pc}) G^T G$

\vspace{0.4em}
\State \textbf{if} $t \pmod T = 0$ \textbf{then} \hfill \Comment{Execute every $T$ steps}
\State \quad $\tilde{L} \leftarrow \frac{\dim(L)}{\text{Tr}(L) + \epsilon} L, \quad \tilde{R} \leftarrow \frac{\dim(R)}{\text{Tr}(R) + \epsilon} R$
\State \quad $Q_L, \Lambda_L \leftarrow \text{Eigh}(\tilde{L} + \epsilon I)$, \quad $Q_R, \Lambda_R \leftarrow \text{Eigh}(\tilde{R} + \epsilon I)$ \Comment{Eigenvectors $Q$}

\State \quad $S_L \leftarrow \Lambda_L^{-\alpha}, \quad S_R \leftarrow \Lambda_R^{-\alpha}$ \Comment{And Eigenvalues $\Lambda$}

\vspace{0.4em}
\Statex \textbf{\textit{Whitening}}
\State $M_{\text{eig}} \leftarrow Q_L^T M Q_R, \quad \tilde{M} \leftarrow S_L M_{\text{eig}} S_R$ 

\vspace{0.4em}
\Statex \textbf{\textit{Spectral Constraint}}
\State $\bar{M} \leftarrow \text{NewtonSchulz}(\tilde{M}), \quad \gamma \leftarrow \|\bar{M}\|_F$ \Comment{Save the norm after NS}

\vspace{0.4em}
\Statex \textbf{\textit{Unwhitening}}
\State $U_{\text{eig}} \leftarrow S_L \bar{M} S_R, \quad U \leftarrow Q_L U_{\text{eig}} Q_R^T$
\State $U \leftarrow \gamma \cdot \frac{U}{\|U\|_F}$ 

\vspace{0.4em}
\Statex \textbf{\textit{Parameter Update}}
\State $\theta \leftarrow \theta - \eta U$
\end{algorithmic}
\end{algorithm}

This result demonstrates that Mousse is equivalent to applying standard spectral optimization within a rectified geometric frame, effectively synthesizing the convergence benefits of Muon with the curvature-awareness of Shampoo. We will discuss this formulation further in Appendix~\ref{appendix:framework} and other details that make this method work in practice in Section~\ref{sec:ablation}.

\subsection{Comparison with Muon and SOAP}

\begin{figure*}[t]
  \includegraphics[width=\columnwidth]{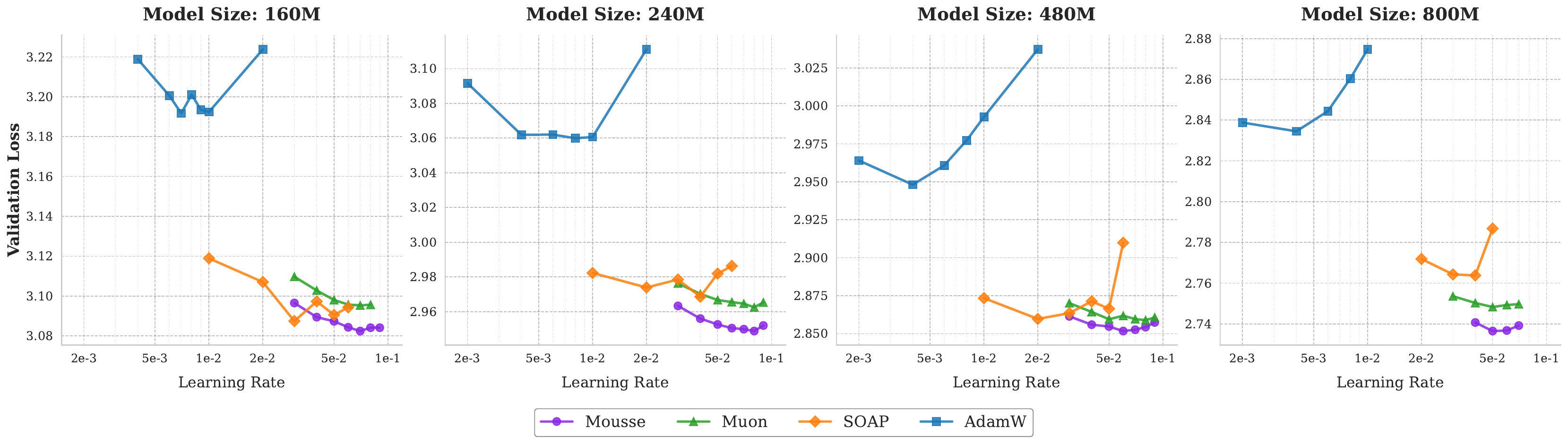}
  \caption{\textbf{Validation Loss comparison on FineWeb (20B tokens).} We report the final validation loss of Mousse against AdamW, Muon and SOAP across varying peak learning rates for model sizes ranging from 160M to 800M. Mousse consistently achieves the lowest validation loss across all model scales, demonstrating superior performance and scalability.}
  \label{fig:main_results}
\end{figure*}

Figure~\ref{fig:comparison} schematically illustrates the structural differences between Muon, SOAP and Mousse. While Muon applies the Newton-Schulz iteration directly to the raw gradient --- implicitly assuming an isotropic landscape --- SOAP explicitly rectifies this geometry by constructing a curvature-aligned eigenbasis derived from preconditioners $L$ and $R$. However, Soap relies on running a complete AdamW optimizer within this rotated frame, necessitating the maintenance of an additional second-momentum states $v$. To rectify Muon's geometric isotropy without such overhead, Mousse employs the same amortized whitening strategy to correct the geometry but strictly adheres to the parameter-efficient Newton-Schulz iteration. Since the spectral constraint inherently normalizes update magnitude, Mousse eliminates the need for the redundant second-momentum state, effectively synthesizing curvature-awareness with the memory efficiency of Muon.

\section{Experiments}

\subsection{Settings}

\textbf{Implementation and Architecture}

We conducted our experiments within a modified version of $\texttt{Dion}$ \citep{ahn2025dion} library, a solid implementation of distributed Muon training. For all experiments, we utilized a standard decoder-only GPT-2 architecture with several modern modifications (Appendix~\ref{appendix:model_structure}). To ensure stable training dynamics across different scales and to facilitate hyperparameter transfer, we adopted the spectral condition~\citep{yang2022tensor} for model initialization and parameter scaling. We evaluated performance across a wide spectrum of model sizes, including 160M, 240M, 480M and 800M parameters.

\noindent \textbf{Training Protocol and Datasets}

To rigorously assess convergence speed and scalability, we trained all models on the $\textbf{FineWeb}$~\citep{penedo2024the} dataset. Each model was trained for 10,000 steps with a global batch size of 2M tokens, amounting to a total of 20 billion tokens. We employed a learning rate schedule consisting of a 10\% linear warmup followed by a cosine decay phase, annealing the learning rate to 0 at the end.

\noindent \textbf{Baselines and Hyperparameter Tuning}

\begin{figure*}[t]
  \includegraphics[width=\columnwidth]{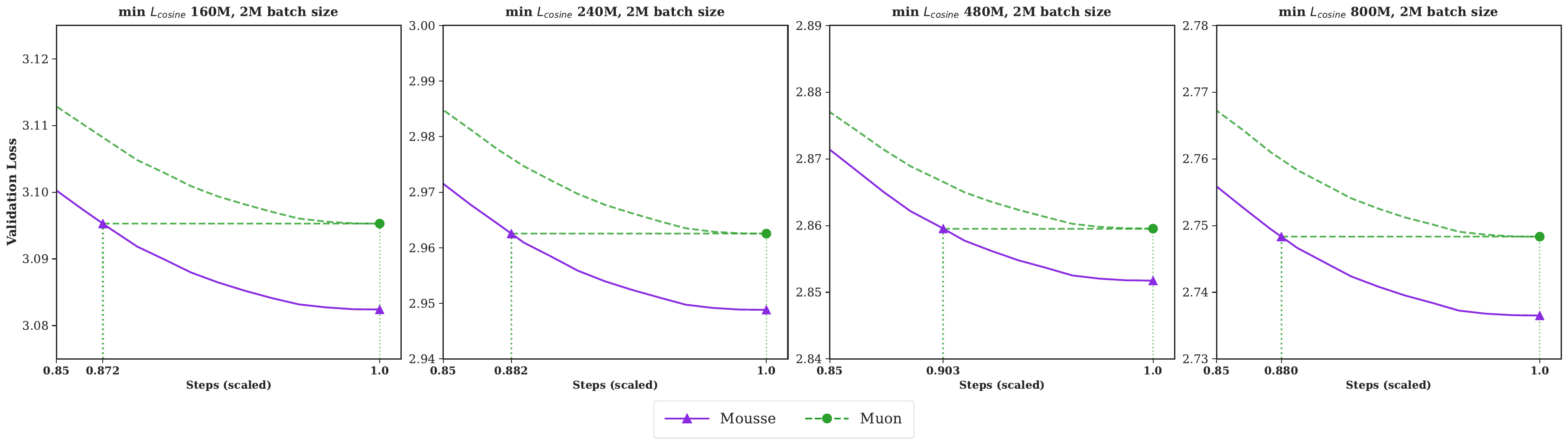}
  \caption{\centering Mousse's performance gains over Muon across 160M, 240M, 480M, and 800M models.}
  \label{fig:muonvsmousse}
\end{figure*}

We compared Mousse against three representative baselines in adaptive and spectral optimization including AdamW, Muon and SOAP.  To ensure a fair comparison, we conducted an extensive grid search for the peak learning rate for every optimizer across all model sizes and settings. Matrix-wise optimizers including Muon, SOAP and Mousse utilize Lion \citep{chen2023symbolic} for embedding and lmhead modules. 

\subsection{Main Results}

Our main results can be categorized into three key advantages:

\paragraph{Superior Convergence Quality.}
As visually summarized in Figure~\ref{fig:main_results}, Mousse demonstrates a strict performance advantage across all model scales. Quantitatively, on the 800M parameter model, Mousse reduces the final validation loss by approximately \textbf{0.012} compared to the best Muon baseline. By integrating Shampoo-style preconditioning, Mousse exploits high-curvature directions effectively, maintaining a distinct lead in the final converged loss. (See more experiment results in Appendix~\ref{appendix:wsd})

\begin{figure}[!b]
  \centering
  \includegraphics[width=0.8 \columnwidth]{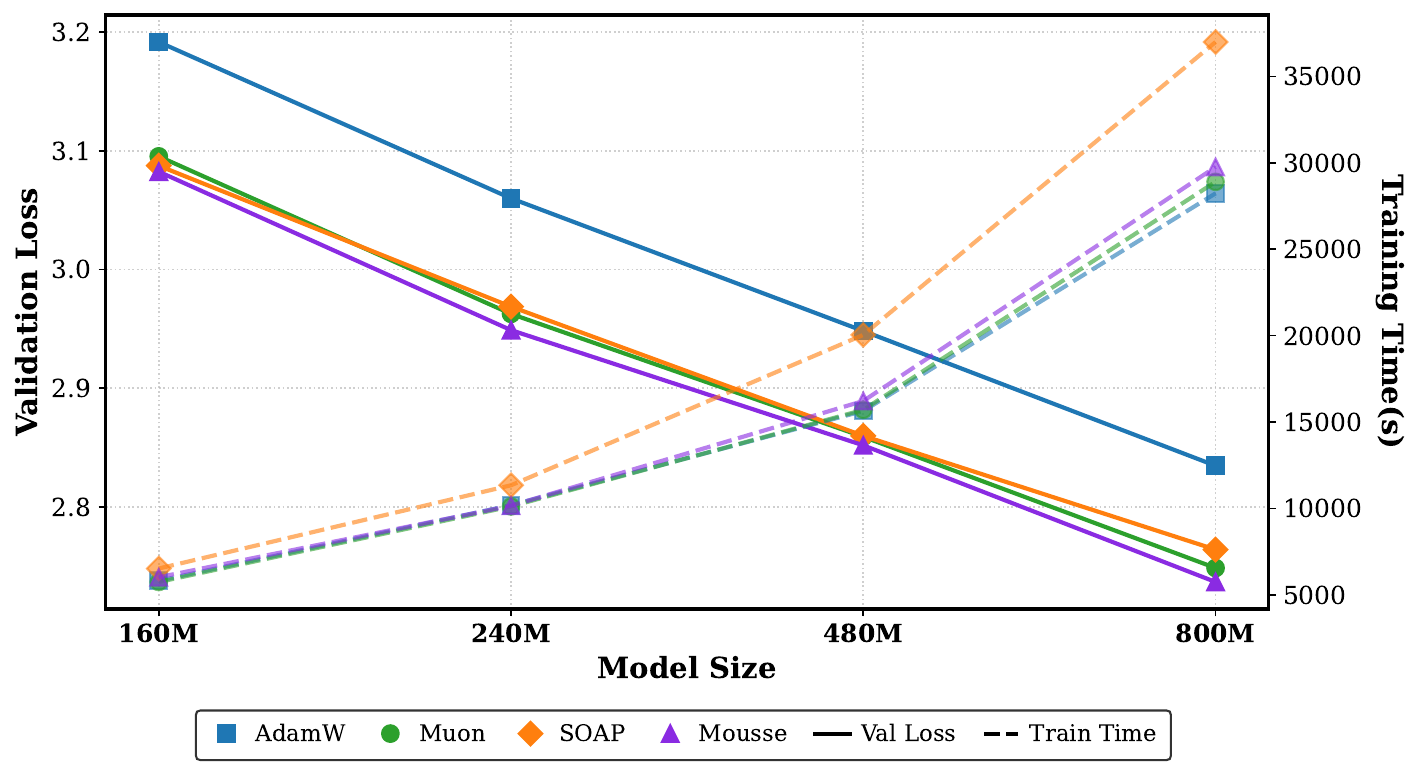}
  \caption{\textbf{Scalability analysis of validation loss and training efficiency.} Solid lines represent validation loss (left y-axis, lower is better), while dashed lines indicate total training time (right y-axis, lower is better) across model sizes ranging from 160M to 800M. Mousse consistently achieves the lowest validation loss across all scales. Crucially, Mousse maintains a training speed nearly identical to the efficient Muon optimizer.}
  \label{fig:moussevssoap}
\end{figure}

An interesting observation from the hyperparameter sweep is that Mousse exhibits a learning rate sensitivity profile remarkably similar to standard Muon, a consequence of their shared spectral optimization foundations. However, unlike Muon, Mousse strictly shifts the performance frontier downwards. This performance gap highlights the limitations of Muon's curvature-agnostic update rule, while Mousse effectively acts as a "geometrically rectified" enhancement of Muon—retaining the stable, wide optimality basin of spectral methods.

\paragraph{Pareto-Optimal Sample Efficiency.}
Another critical advantage of Mousse lies in its sample efficiency. Mousse exhibits a significantly steeper descent trajectory once the curvature statistics stabilize. Quantitatively (see Figure~\ref{fig:muonvsmousse}), Mousse requires approximately \textbf{12\% fewer training steps} to reach the final converged loss level of the Muon baseline. Figure~\ref{fig:moussevssoap} visualizes this scalability trade-off across model sizes. The dashed lines (Total Training Time) reveal that Mousse maintains a training speed nearly identical to the highly efficient Muon optimizer, with negligible overhead. In sharp contrast, SOAP suffers from significant throughput degradation. By achieving strictly lower validation loss without the computational penalty typical of curvature-corrected methods, Mousse effectively establishes a new Pareto frontier for large-scale pre-training efficiency.

\paragraph{High Throughput with Low Overhead.}
Despite incorporating second-order information, Mousse incurs less computational and memory overhead than traditional methods like SOAP. By relying on the spectral update via Newton-Schulz iterations, Mousse intrinsically regulates step size, eliminating the need for a full-sized FP32 variance buffer required by Adam-style optimizers. Furthermore, when combined with the single-sided preconditioning technique detailed in Section~\ref{sec:single_sided}, Mousse's memory footprint is further reduced, bringing peak memory usage to approximately \textbf{88\%} of that of SOAP. This is comparable to the lightweight Muon optimizer (about \textbf{1.05x} that of Muon), while still delivering superior convergence. Table~\ref{tab:optim_comparison} shows an overall comparison between different optimizers.

\begin{table}[h]
    \centering
    \caption{\textbf{Optimizer Comparison.} We evaluate AdamW, Muon, SOAP, and our proposed Mousse across training speed, memory efficiency, and final performance. ``$++$'' indicates the best results.}
    \label{tab:optim_comparison}

    \resizebox{0.5\linewidth}{!}{
        \begin{tabular}{lcccc}
            \toprule
            \textbf{Metric} & AdamW & Muon & SOAP & \textbf{Mousse (Ours)} \\
            \midrule
            Training Speed & $++$ & $++$ & $-$ & $\mathbf{++}$ \\
            Memory Efficiency & $++$ & $++$ & $-$ & $\mathbf{+}$ \\
            Final Performance & $-$ & $+$ & $+$ & $\mathbf{++}$ \\
            \bottomrule
        \end{tabular}
    }
\end{table}

\section{Ablation \& Analysis}
\label{sec:ablation}

\subsection{Grafting}

Gradient grafting is a technique that decouples the optimization into a robust magnitude derived from a stable method like AdamW and another superior direction from a more sophisticated optimizer like Shampoo, which has been identified as "the key ingredient to making Shampoo work in practice" \citep{shi2023pytorchshampoo}. We conduct an ablation study to determine whether explicitly controlling the update norm benefits Mousse. Theoretically, Mousse does not strictly require an auxiliary optimizer for magnitude control, as it can directly utilize the intrinsic unit spectral norm of the orthogonalized update as a stable step size.

\begin{figure*}[t]
  \includegraphics[width=0.48\linewidth]{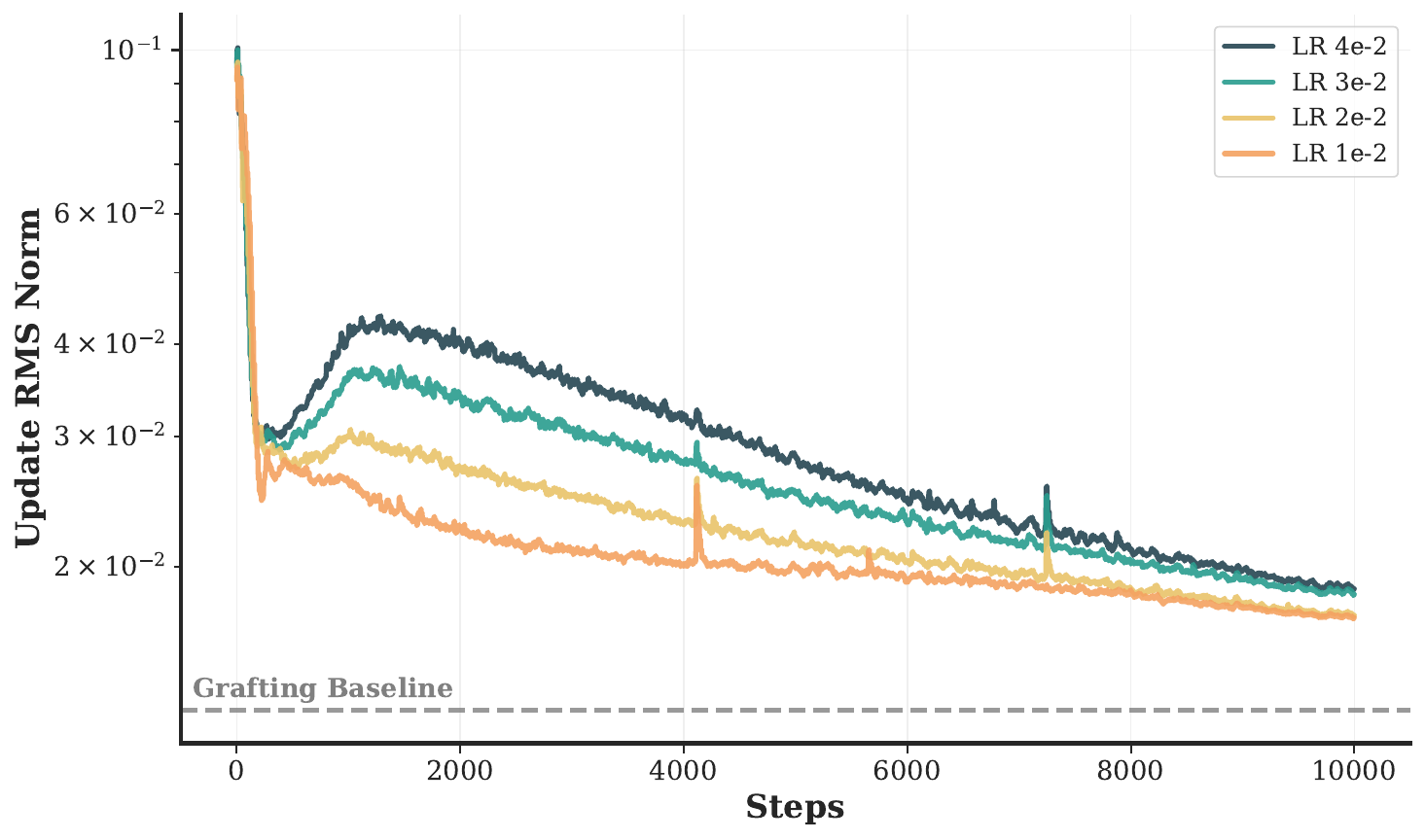} \hfill
  \includegraphics[width=0.48\linewidth]{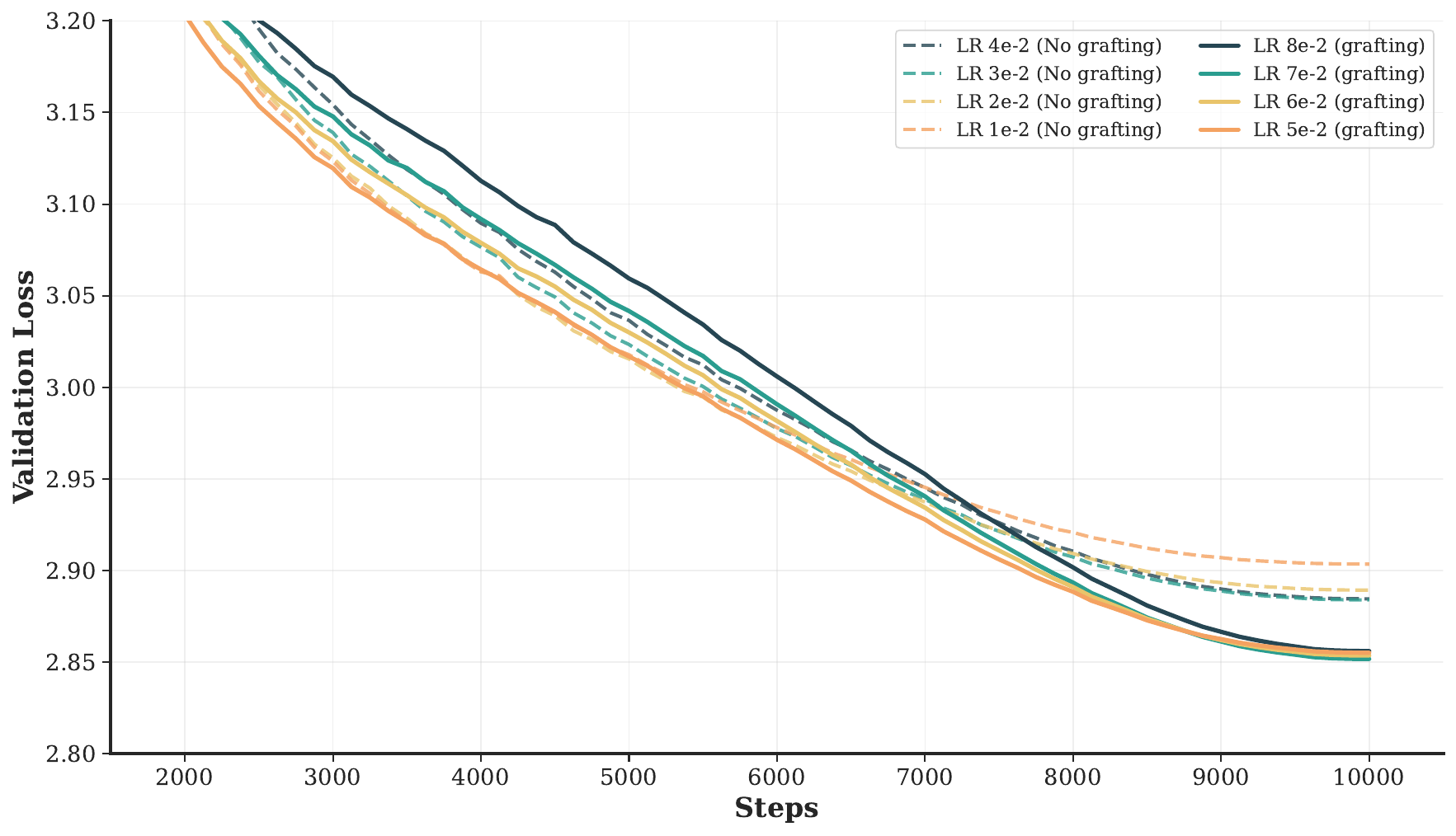}
  \caption {\textbf{Impact of Gradient Grafting on Update Stability.} \textbf{(Left) RMS Norm:} We visualize the RMS norm of parameter updates throughout the training trajectory across varying learning rates. Solid lines represent settings without grafting, exhibiting a continuous downward drift independent of the learning rate schedule. The dashed line represents the constant update norm constrained by spectral normalization when grafting is enabled. \textbf{(Right)Validation Loss} trajectories for both settings around their respective optimal learning rates: By decoupling the magnitude from the optimization direction, grafting ensures superior convergence rates and lower final validation loss.}
  \label{Fig:grafting}
\end{figure*}

As shown in Figure~\ref{Fig:grafting}, without external grafting, we observe a continuous downward drift in the update RMS norm irrespective of the cosine learning rate scheduler. 
These diminishing updates not only impede convergence speed but also introduce unnecessary complexity to learning rate tuning. However, applying grafting effectively counteracts this trend by maintaining a stable update magnitude. Our empirical results confirm that the ungrafted baseline exhibits performance degradation in later training stages, attributed to insufficient or unstable updates from the decaying RMS norm. In contrast, incorporating grafting consistently yields superior performance and stability.

\subsection{Condition Control}

The stability of Mousse hinges on the well-conditioned decomposition of the Hessian approximations $L$ and $R$. A critical challenge arises from the computation of the negative fractional power ($L^{-\alpha}$). In the eigenspectrum of deep neural networks, "flat" directions correspond to small eigenvalues $\lambda_i \approx 0$. When raised to a negative power, these directions result in large scaling factors, risking catastrophic gradient amplification along noisy, low-curvature axes. We compared several different solutions for addressing stability issues, shown in Figure~\ref{fig:trace_normalization}.

\paragraph{Trace Normalization.} Empirically, we observe that the absolute magnitudes of $L$ and $R$ vary significantly across layers, often decaying to small numerical values after the initial gradient norm convergence (Appendix~\ref{appendix:preconditioner_stability}). This scale variance makes it difficult to apply a uniform damping hyperparameter $\epsilon$. To adapt to this, we propose \textbf{Trace Normalization}. Before decomposition, we normalize the covariance matrices such that the mean eigenvalue is unity (i.e., $\text{Tr}(L) = \text{dim}(L)$ and $\text{Tr}(R) = \text{dim}(R)$). This ensures that the damping term $\epsilon$ has a consistent relative effect across all modules.

\paragraph{Spectral Tempering.} Empirically, we investigated the sensitivity of Mousse to the "strength" of the curvature correction, governed by the exponent $\alpha$ in the whitening factors $L^{-\alpha}$ and $R^{-\alpha}$ and the damping factor $\epsilon$. While standard Shampoo theoretics suggest $\alpha = 0.25$, we found that Mousse cannot endure such aggressive curvature correction.

Our experiments reveal that a milder exponent of $\alpha = 0.125$ consistently outperforms the standard $\alpha=0.25$, which superimposes a sharp full-strength curvature correction distorting the update direction. A larger damping term also helps moderate the curvature strength, preventing excessive step sizes in flat directions. This "Spectral Tempering" strategy strikes an optimal balance between the isotropic stability of spectral optimization and the anisotropic acceleration of second-order methods.

\subsection{Single-Sided Preconditioner}
\label{sec:single_sided}

Inspired by ASGO~\citep{an2025asgo, cesista2025sdnr}, we investigate a variant of Mousse that employs a single-sided preconditioner substituting the full Kronecker product $R \otimes L$ with only one factor. Crucially, this approximation halves both the computational cost of the eigen-decomposition and the memory footprint of the preconditioner states.

\begin{figure}[t]
  \centering
  \begin{subfigure}[b]{0.5\columnwidth}
    \centering
    \includegraphics[width=\linewidth]{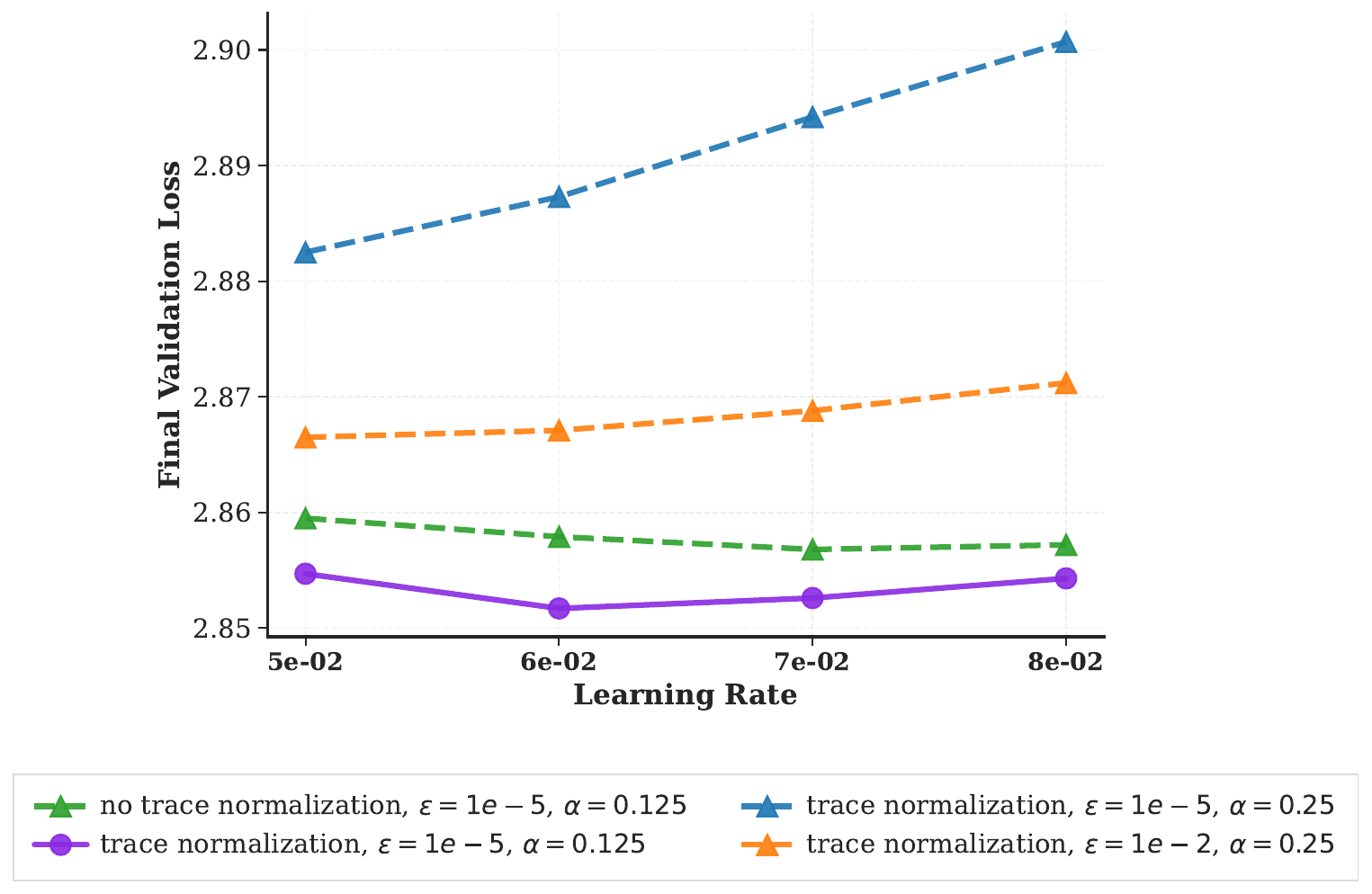}
    \caption{\centering Condition Control Strategies}
    \label{fig:trace_normalization}
  \end{subfigure}
  \hfill
  \begin{subfigure}[b]{0.44\columnwidth}
    \centering
    \includegraphics[width=\linewidth]{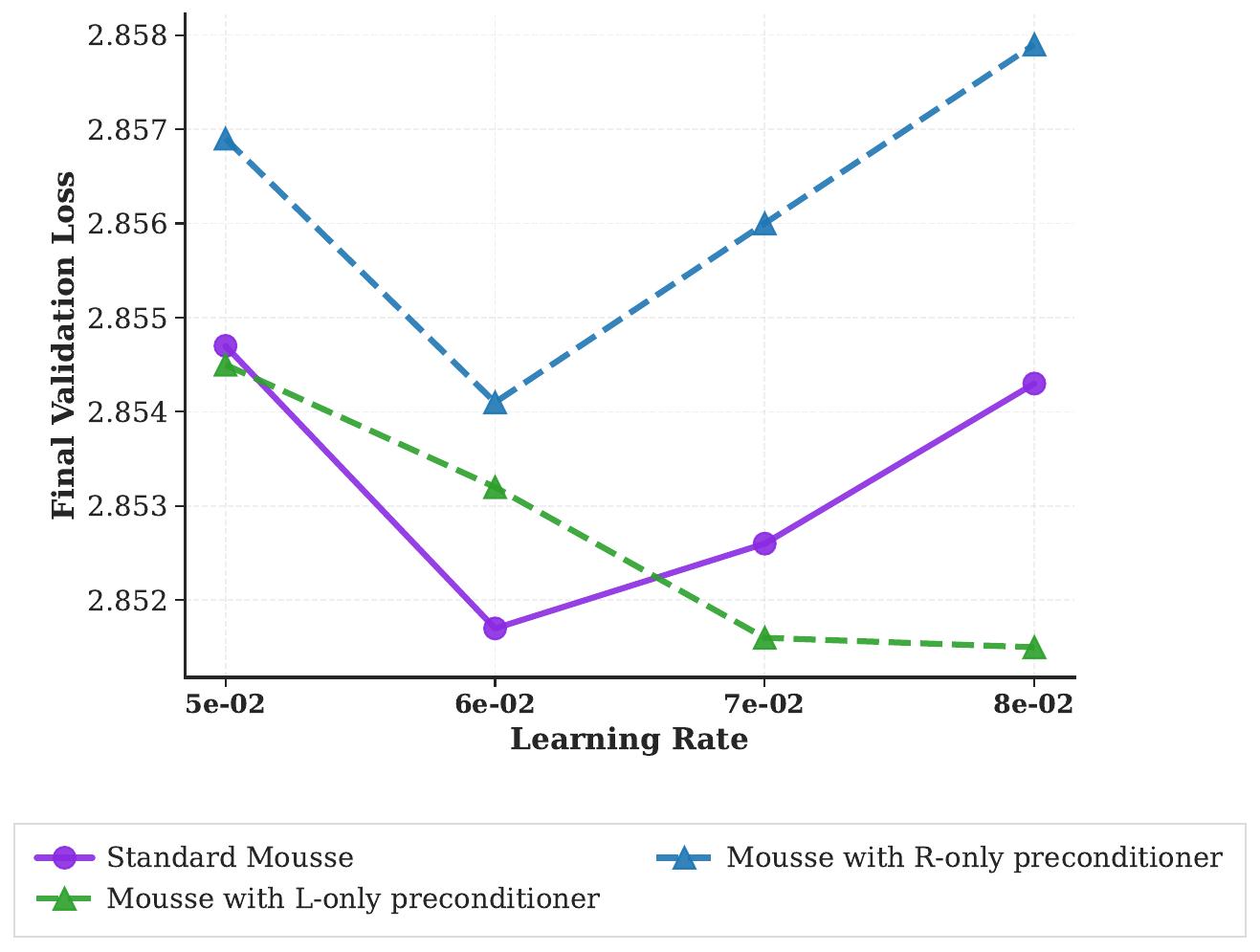}
    \caption{\centering Single vs. Double-sided Preconditioner}
    \label{fig:oneside}
  \end{subfigure}

  \caption{\textbf{Ablation studies on 480M model.} \textbf{(a)} We compare the validation loss trajectories under different preconditioning settings. The results demonstrate that \textbf{Trace Normalization} is essential for stability. Furthermore, \textbf{Spectral Tempering} ($\alpha=0.125$) consistently outperforms the aggressive curvature correction correction ($\alpha=0.25$), yielding the lowest final loss. \textbf{(b)} Comparison of single-sided and double-sided preconditioners.}
  \label{fig:combined_ablation}
\end{figure}

Our experiments (Figure~\ref{fig:oneside}) demonstrate that this method achieves comparable performance to the Mousse baseline, yielding a negligible decline or even slight improvements. This empirical finding underscores the feasibility of the single-sided whitening.

Interestingly, we observed that using the left-sided preconditioner ($L$) is consistently slightly better than the right-sided preconditioner ($R$). We hypothesize that this advantage stems from the presence of the preceding LayerNorm, which typically standardizes the input activations (whose statistics are captured by $L$). This regularization renders the curvature information in the activation axis more reliable and critical for stable optimization than that of the gradient axis.

\section{Future Works}

Here we list several promising directions that could either enhance the performance of Mousse or elevate its capabilities but haven't been fully discussed in this work yet due to time and computation constraints. 

\subsection{Better Shampoo Preconditioner}

A key strength of Mousse is its inherent compatibility with the rapid improvements in preconditioner design. Following our exploration of single-sided preconditioning in Section~\ref{sec:single_sided}, we foresee significant potential in integrating other recent techniques. For example, \citet{duvvuri2024combining} proposed to use Kronecker-Sum Preconditioner instead of Shampoo's original factor for better stability. \citet{eschenhagen2025purifying} also pointed that different modules should have different eigen base update frequency, aiming at saving computation resources and for better performance. These strategies may further enhance Mousse's ability.

\subsection{Better Engineering Practices}

Our current implementation of Mousse represents a proof-of-concept realization, focusing primarily on algorithmic correctness rather than hardware efficiency. Consequently, there remains significant headroom for system-level optimizations, such as adopting power iteration with QR decomposition or Newton-Schulz iteration for spectral decomposition~\citep{shi2023pytorchshampoo} instead of naive \texttt{torch.linalg.eigh}.

\subsection{Bridging the Gap in Fine-tuning}

Standard Muon often struggles to fine-tune models pre-trained with AdamW\citep{liu2025muon}. We hypothesize that as Mousse explicitly incorporates second-order information, it shares a similar adaptive nature with Adam-style methods, thus offering smoother transition for fine-tuning.

\section{Conclusion}

In this work, we introduce \textbf{Mousse}, an optimizer that enhances spectral optimization by explicitly integrating structural curvature information. By performing Newton-Schulz orthogonalization within a whitened coordinate system, Mousse aligns the update step with the anisotropic geometry of the loss landscape. Empirical results on language models up to 800M parameters demonstrate that Mousse consistently yields lower validation loss compared to Muon, with negligible computational overhead. Finally, our analysis of stability techniques, including Trace Normalization and Spectral Tempering, establishes a robust framework for second-order spectral optimization, positioning Mousse as a promising solution for large-scale pre-training.

\section{Acknowledgments}

This research project was supported by Shanghai Artificial Intelligence Laboratory. We further extend our gratitude to Tianyang Lin and  Runyu Peng for their insightful feedback and discussions with us.

\bibliographystyle{plainnat}
\bibliography{custom}

\appendix

\section{Appendix}
\label{sec:appendixA}

\subsection{General Framework of Geometric Spectral Descent}
\label{appendix:framework}

We have mentioned in~\ref{sec:mousse} that different approximations of $H$ yield different algorithms under the framework of geometric spectral descent. To be specific, let's denote the preconditioner as $\tau_H(\cdot)$, and the target problem can be written as:
\begin{align*}
    \Delta W = \underset{U}{\arg\min} \langle G, U \rangle \quad \\
    \text{s.t.} \quad \|\tau_H(U)\|_{\text{op}} \leq 1
\end{align*}

If we assume that $H$ is approximated by Adam's second momentum $\sqrt{v} = \sqrt{g^2}$, then the whitening form becomes:
\begin{align*}
    \Delta W = \underset{U}{\arg\min} \langle G, U \rangle \quad \text{s.t.} \quad \|\sqrt{v} \odot U\|_{\text{op}} \leq 1
\end{align*}

Similarly, by performing a change of variables $Y = \sqrt{v} \odot U$, which implies $U = \frac{Y}{\sqrt{v}}$, the objective becomes:
\begin{align*}
    &\langle G, \frac{Y}{\sqrt{v}} \rangle = \langle \frac{G}{\sqrt{v}}, Y \rangle \\
    \Rightarrow &Y^* = -\text{msign}(\frac{G}{\sqrt{v}}) \\
    \Rightarrow &\Delta W = - \frac{1}{\sqrt{v}} \odot \text{msign}(\frac{G}{\sqrt{v}})
\end{align*}

Interestingly, a recent variant termed AdaMuon\citep{si2025adamuon} proposes a similar update idea however diverges in how the second momentum is accumulated. Specifically, AdaMuon updates the scaling factor $\sqrt{v}$ using the orthogonalized update $O = \text{msign}(M)$, rather than the raw gradient or momentum. From our perspective, this does not fit our geometric framework theoretically, and its empirical success might instead stem from an implicit regularization effect or stabilized magnitude control, rather than genuine curvature-aware optimization.

Moreover, FISMO\citep{xu2026fismo} independently develops a similar geometric perspective by adopting a KFAC-based block-diagonal approximation of $H$. However, FISMO's reliance on the \textbf{empirical Fisher} has been subject to a long-standing criticism\citep{kunstner2019limitations} because of its skewed distribution compared with the real Fisher information matrix. Furthermore, FISMO suffers from a fatal flaw that maintaining its iterative K-FAC preconditioner updates requires recomputing the inverse matrices at \textbf{every single step}. This imposes a prohibitive computation cost, rendering it practically infeasible for large-scale training where wall-clock efficiency is paramount. In sharp contrast, Mousse adopts an \textbf{amortized update strategy} (similar to Shampoo and SOAP), recomputing the preconditioner only every $T$ steps (e.g. $T=10$ in our experiments). This drastically reduces the overhead without compromising convergence stability.

\subsection{Natural Pairing}
\label{appendix:natural_pairing}

In Section~\ref{sec:mousse}, we utilize the notation $\langle G, \Delta W \rangle$ to formulate the optimization objective. It is crucial to clarify that strictly speaking, this denotes the \textbf{natural pairing} between a vector space and its dual, rather than a geometric inner product induced by a metric~\citep{absil2008optimization}.

Mathematically, the update step $\Delta W$ resides in the tangent space $\mathcal{T}\mathcal{W}$, while the gradient $G$ naturally resides in the cotangent (dual) space $\mathcal{T}^*\mathcal{W}$ as a linear functional. The operation $\langle G, \Delta W \rangle$ represents the action of the functional $G$ on the vector $\Delta W$, measuring the directional derivative of the loss.

When the parameter space is instantiated as matrices $\mathbb{R}^{m \times n}$, this abstract evaluation concretely manifests as the component-wise sum of products. Algebraically, this is computed via the trace operator:
\begin{equation}
    \langle G, \Delta W \rangle := \sum_{i,j} G_{ij} \Delta W_{ij} = \text{Tr}(G^T \Delta W)
\end{equation}

\section{Supplementary Empirical Results}

\subsection{Loss Performance on WSD scheduler}
\label{appendix:wsd}

\begin{figure}[h]
  \includegraphics[width=\columnwidth]{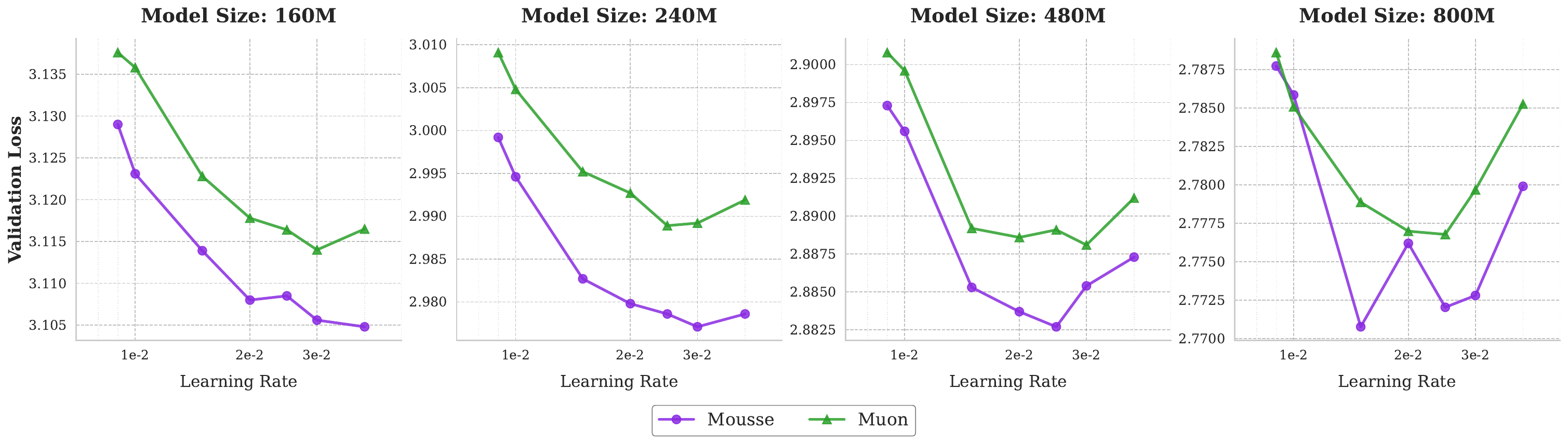}
  \caption{\textbf{Validation Loss comparison on FineWeb (20B tokens) under the WSD scheduler.} We report the final validation across varying peak learning rates for model sizes ranging from 160M to 800M.}
  \label{fig:main_exp_WSD}
\end{figure}

Besides the standard practice of cosine learning rate scheduler, the Warmup-Stable-Decay (WSD) schedule had emerged as a perferred choice in large-scale pre-training practice due to its flexibility for continuous training. Hence, we assess the robustness of Mousse under this regime against Muon using the WSD scheduler. We inherit the same experimental setup with our main experiment, and configure a 10\%/10\% warmup/warmdown phase. The results are showed in Figure~\ref{fig:main_exp_WSD}, where Mousse still consistently outperforms Muon in all settings, showcasing its great robustness.

\subsection{Excessive Loss}
\label{appendix:excessive_loss}

\begin{figure}[h]
\centering
  \includegraphics[width=0.8 \columnwidth]{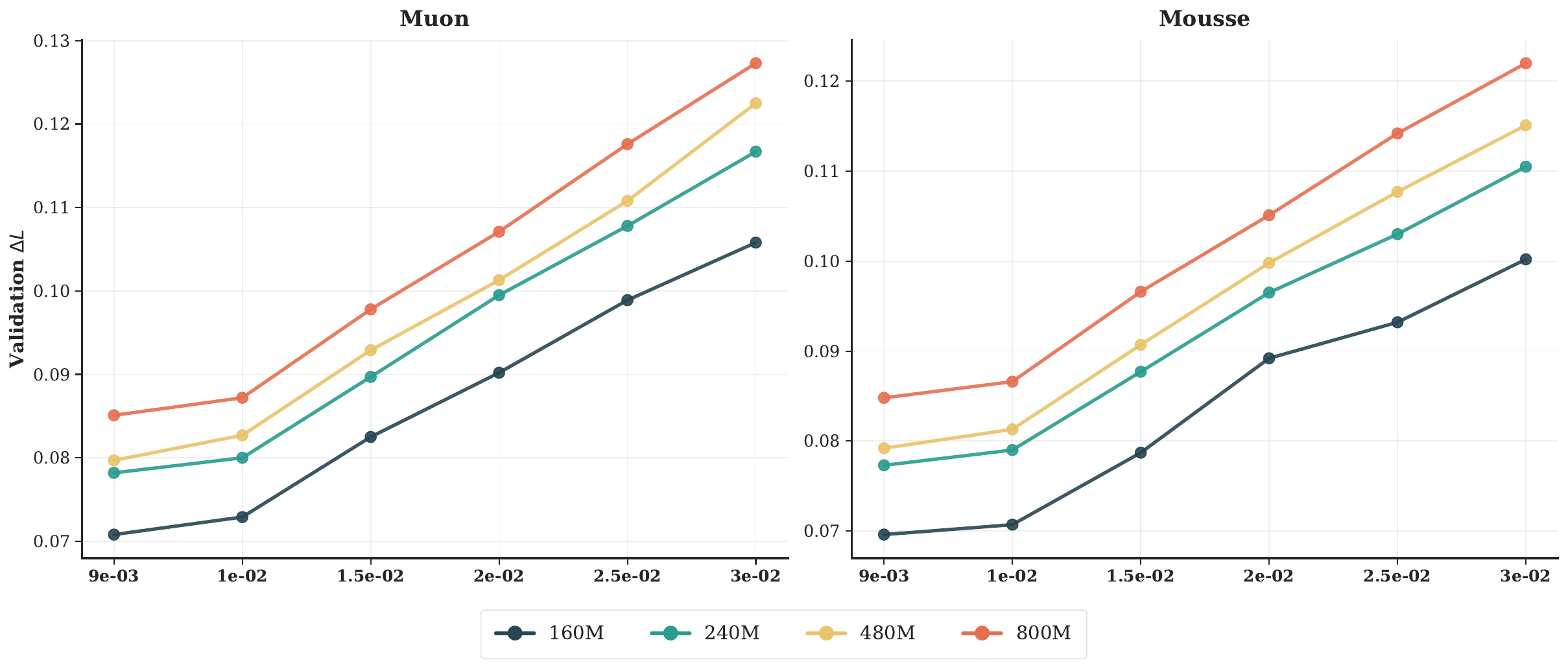}
  \caption{\textbf{Quantification of Excessive Loss} for WSD scheduler. We plot the magnitude of the excessive loss ($\Delta L = L_{stable} - L_{decay}$) during the learning rate decay phase against the peak learning rate for AdamW, Muon, SOAP and Mousse across various model scales. A consistent positive correlation is observed: higher learning rates sustain a higher "noise floor" (excessive loss) during the stable phase, which is subsequently released upon decay.}
  \label{fig:excessive_loss}
\end{figure}

We would like to highlight a critical yet often underappreciated factor that biases fair performance comparisons we called \textbf{excessive loss}. Comparing models based on the validation loss during the stable learning rate phase can be misleading, because the loss value before the learning rate decay schedule does not strictly reflect the model's convergence quality. As illustrated in Figure~\ref{fig:excessive_loss}, the excessive loss ($\Delta L = L_{stable} - L_{decay}$) is positively correlated with both the learning rate and the model scale. 

Previous work\citep{yaida2018fluctuation} has theoretically characterized this phenomenon, noting that the stationary distribution of SGD iterates maintains a variance related to the learning rate and the noise covariance. At the same time, there is no evidence that excessive loss values are comparable across different methods, given their distinct internal noise structures. Consequently, we argue that $L_{\text{decay}}$ serves as a more accurate indicator of the effectiveness of optimizer in finding the global minimum. Otherwise we will disadvantage curves of larger learning rates, as their higher noise floor obscures the true depth of the basin found.

\subsection{The Stability of Preconditioner}
\label{appendix:preconditioner_stability}

\begin{figure}[h]
\centering
  \includegraphics[width=0.95 \columnwidth]{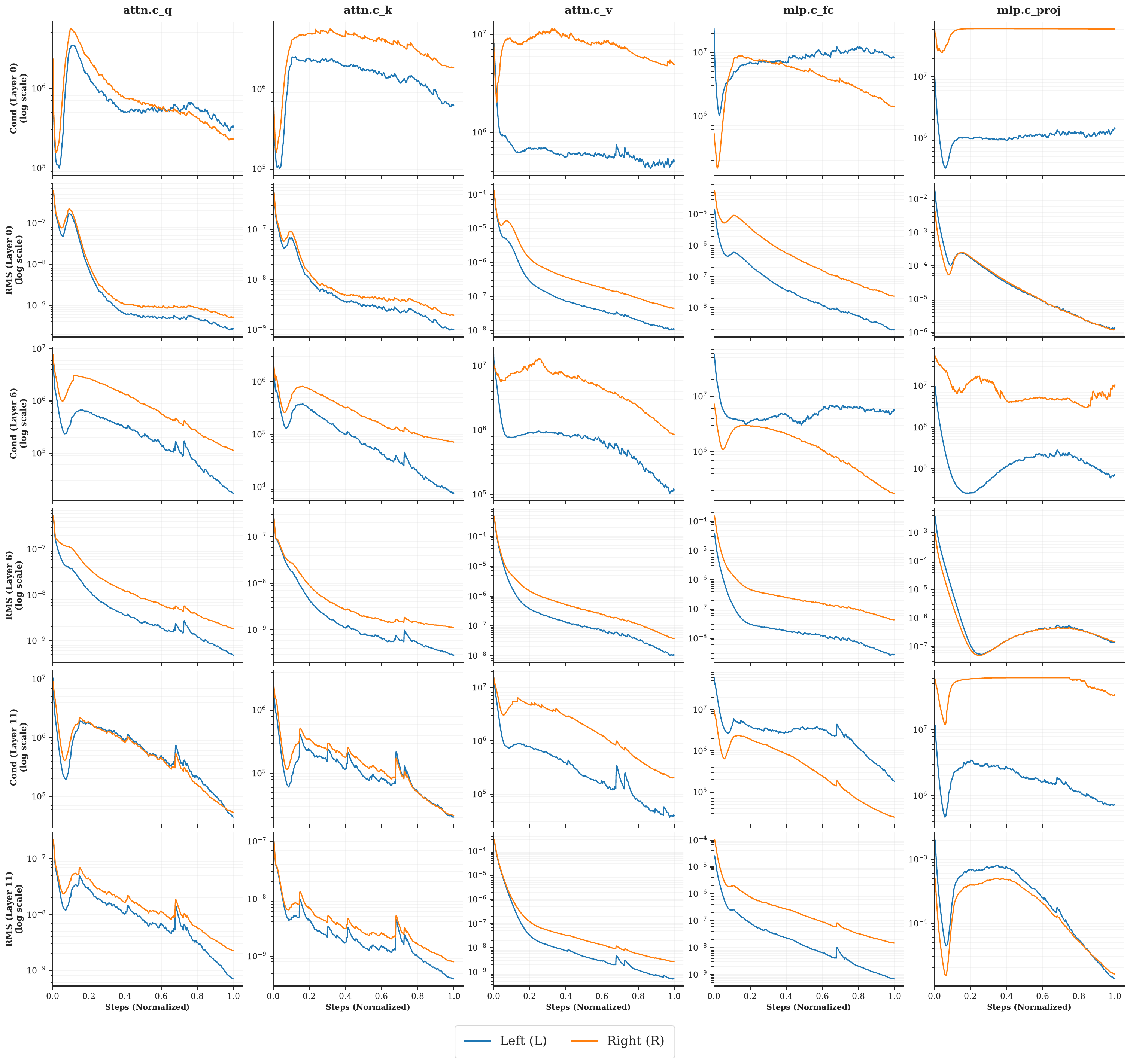}
  \caption{\textbf{Evolution of Preconditioner Stability} on 480M model trained by Mousse with cosine LR scheduler. We visualize the condition numbers and RMS norms of $L$ and $R$ across different network layers during training. Under the trace normalization, most curves showed a relatively stable downward trend, preventing numerical divergence. The preconditioner statistics are accumulated via an Exponential Moving Average (EMA) with $\beta=0.95$.}
  \label{fig:preconditioner_stability}
\end{figure}

We analyze the dynamics of the preconditioner matrices $L$ and $R$ throughout the training process. As illustrated in Figure~\ref{fig:preconditioner_stability}, the raw spectral magnitudes of these matrices exhibit drastic variations across different layers and architectural modules. Furthermore, their temporal fluctuations are non-uniform: some layers experience rapid covariance growth while others remain relatively static or decay.

This heterogeneity necessitates a unified trace normalization strategy. Without it, a fixed damping factor $\epsilon$ would act inconsistently --- negligible for layers with large norms yet dominating for those with small norms, leading to unstable whitening. we observe that the condition numbers of $L$ and $R$ are high during the initial statistic accumulation phase but quickly plateau into a stable range. This indicates that the geometry of the whitened space stabilizes early in training, providing a robust and consistent foundation for the subsequent Newton-Schulz iterations. However, we do observe occasional spikes in the condition numbers within specific modules, suggesting localized instability or sudden shifts in the gradient landscape. This hints that a more adaptive or per-layer damping strategy could further mitigate these transient fluctuations and enhance robustness, which we leave for future exploration.

\section{Details of Experiments}

\subsection{Pretraining GPT2}
\label{appendix:model_structure}

While we follow the general blueprint of the decoder-only GPT-2, we incorporate several modern architectural improvements to ensure training stability and efficiency. Our implementation is based on the $\texttt{dion}$ library \citep{ahn2025dion}. The specific settings include:

\paragraph{Normalization.} We utilize RMSNorm\citep{zhang2019root} before sublayers instead of standard LayerNorm for better stability. To mitigate attention logit instability at scale, we apply QK-Norm\citep{henry2020query}, normalizing the query and key vectors prior to the dot-product attention.

\paragraph{Positional Embeddings.} We use Rotary Positional Embeddings (RoPE)\citep{su2024roformer}. 

\paragraph{Activation Function.} We employ the squared ReLU activation ($f(x) = \text{ReLU}(x)^2$)\citep{so2021searching}, which has been shown to offer superior performance compared to GeLU\citep{hendrycks2016gaussian}. 

\paragraph{Bias.} We disable bias terms in all linear layers and layer normalizations to improve memory efficiency and training stability.

\paragraph{Initialization.} We adopt a variance-scaling initialization scheme designed to facilitate hyperparameter transfer, following the principles of \textbf{Spectral Condition} \citep{yang2023spectral}. specifically, weights are initialized from a normal distribution $\mathcal{N}(0, \sigma^2)$, where the standard deviation $\sigma$ is determined by the fan-in and fan-out dimensions:
\begin{equation}
    \sigma = \frac{1}{\sqrt{\text{fan\_in}}} \cdot \min\left(1, \sqrt{\frac{\text{fan\_out}}{\text{fan\_in}}}\right)
\end{equation}

Embedding weights are initialized with $\mathcal{N}(0, 1)$. This initialization ensures that activation variances remain stable across widths.

\paragraph{Model Configurations.} We trained models across varying scales ranging from 160M to 800M parameters. The specific configurations are detailed in Table~\ref{tab:model_configs}. All models utilize a context window of 1024 tokens and a vocabulary size of 50,304 (padded from 50,257 for kernel efficiency).

\begin{table}[h]
    \centering
    \caption{\centering Precise Model Architecture Configurations. All models use the architecture described above.}
    \label{tab:model_configs}
    \renewcommand{\arraystretch}{1.1}
    \resizebox{0.5 \columnwidth}{!}{
        \begin{tabular}{lcccc}
            \toprule
            \textbf{Model} & \textbf{Embedding} & \boldmath$d_{\text{model}}$ & \boldmath$n_{\text{layers}}$ & \boldmath$n_{\text{heads}}$ \\
            \midrule
            162M & 38M & 768 & 12 & 6 \\
            247M & 38M & 768 & 24 & 6 \\
            494M & 77M & 1536 & 12 & 6 \\
            834M & 77M & 1536 & 24 & 6 \\
            \bottomrule
        \end{tabular}
    }
\end{table}

\paragraph{Optimizer Settings.}

\begin{itemize}
    \item \textbf{AdamW:} We use $\beta_1=0.9, \beta_2=0.95$, $\epsilon=1e-8$, and a weight decay of 0.01.
    \item \textbf{Muon:} We use $\beta_1=0.95$, a weight decay of 0.01, and the standard configuration with the NS5 kernel.
    \item \textbf{SOAP:} We use $\beta_1=0.95, \beta_2=0.95, \epsilon=1e-8$, and a weight decay of 0.01. The preconditioning update frequency is set to $T=10$. The whitening statistics ($L, R$) are accumulated using an EMA with $\beta_{pc}=0.95$.
    \item \textbf{Mousse:} We align the optimization hyperparameters ($\beta_1=0.95$, weight decay of 0.01) and the Newton-Schulz kernel (NS5) with Muon. Additionally, we adopt the preconditioning update frequency ($T=10$) and the whitening statistics EMA ($\beta_{pc}=0.95$) from SOAP. We set $\epsilon=1e-5$ for spectral decomposition.
\end{itemize}

\paragraph{Training.} All models were trained using Distributed Data Parallel (DDP) on a cluster of 8 Nvidia H200 GPUs. Models are trained for 10,000 steps with a global batch size of 2 Million tokens. The learning rate schedule follows a linear warmup for the first 1,000 steps (10\%), followed by a cosine annealing to 0 until the final step.


\end{document}